# Recognition of Images of Korean Characters Using Embedded Networks

Sergey A. Ilyuhin[1,3], Alexander V. Sheshkus[1,2], Vladimir L. Arlazarov[2,3]
[1]Smart Engines Service LLC;
[2]Federal Research Center "Computer Science and Control" of RAS, Moscow, Russia;
[3]Moscow Institute of Physics and Technology, Moscow, Russia;

## ABSTRACT

Despite the significant success in the field of text recognition, complex and unsolved problems still exist in this field. In recent years, the recognition accuracy of the English language has greatly increased, while the problem of recognition of hieroglyphs has received much less attention. Hieroglyph recognition or image recognition with Korean, Japanese or Chinese characters have differences from the traditional text recognition task.

This article discusses the main differences between hieroglyph languages and the Latin alphabet in the context of image recognition. A light-weight method for recognizing images of the hieroglyphs is proposed and tested on a public dataset of Korean hieroglyph images. Despite the existing solutions, the proposed method is suitable for mobile devices. Its recognition accuracy is better than the accuracy of the open-source OCR framework. The presented method of training embedded net bases on the similarities in the recognition data.

## 1. INTRODUCTION

Over the years, optical character recognition has been a popular research topic for computer vision specialists [1-6]. Convolutional neural networks have proven themselves as a good solution for such problems as object recognition. For those networks, the number of outputs is equal to the number of characters of the recognizing alphabet. Such networks are small and show high speed and recognition accuracy, for example, for the English language. However, this method has difficulties in solving the same problem for the Korean language.

Hieroglyphs considerably differ from letters, for example, Korean hieroglyphs are always constructed from simpler glyphs or keys. There are three groups of keys and a hieroglyph can be built from two or three keys from the different groups. While keys from the first two groups are required in the hieroglyph, the key from the third group is optional. There are 19, 21 and 27 keys in the first, second and third group respectively. Therefore, there can be 10773 hieroglyphs constructed from three keys and 399 constructed from two keys are possible. That yields to 11172 different hieroglyphs in total.

The number of Korean hieroglyphs is 432 times larger than the number of characters in the English alphabet (26 against 11172). Therefore, classification neural network for hieroglyphs will have a huge amount of trainable parameters and consequently will be hard in training and time-consuming in usage. Despite a big number of characters it also exists an issue about the complexity of some hieroglyphs Fig. 1. Many hieroglyphs consist of small parts that could be easily detected as a noise on the image or a part of the background. Such cases force developers and researchers to use only high resolution and high scale images for recognition.

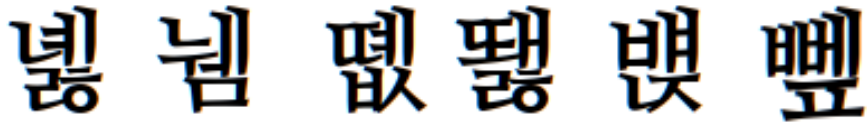

Figure 1. Example of complexity of hieroglyphs.

Another issue is the similarity of some hieroglyphs Fig 2. For these hieroglyphs, some minor changes could transform one symbol to another one. This is why the background or some distortions that were made while taking the picture of these symbols can confuse the classifier. In the context of training of classifier it problem put a limitation on the augmentation of the training data.

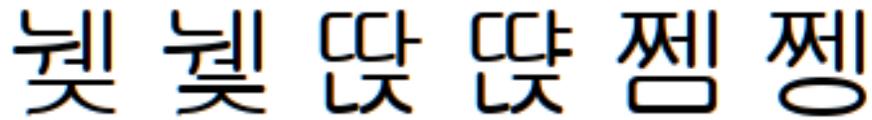

Figure 2. Example of similarity of different hieroglyphs.

In omnifont hieroglyph recognition another problem arises. The appearance of a specific symbol can vary considerably from font to font as shown in Fig. 3. One more problem for hieroglyphs is different in writing one symbol through different fonts Fig 3. This problem is valid for all the languages but with hieroglyphs, the same symbol from different fonts can vary more than different symbols from one font. For example, the first and the last symbols in Fig. 3 are less similar than the third and fourth symbols in Fig. 2. Therefore, we need to know a font or have additional information to give a correct answer in some cases.

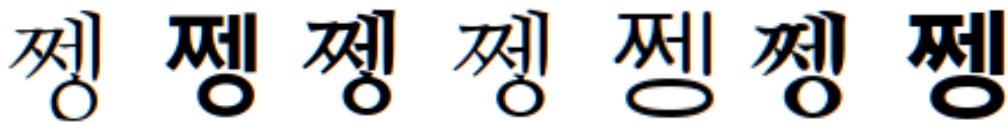

Figure 3. Example of one Korean symbol through different fonts.

Due to the large number of characters in the Korean language alphabet and their variability, it is necessary to enlarge the size of the classification network by increasing the number of outputs in the last layer, which complicates its training and increases the processing time for a one symbol image.

This article discusses the use of one of the alternative classification methods using the architecture of Siamese networks [7] to train an embedded net, which is used in the face recognition problem [8].

## 2. RELATED WORK

### 2.1 Hieroglyphs Recognition

Image recognition solves the classification problem: it associates images with classes to which they belong. The recognition quality of English text is at a very high-level [9]. Since the size of hieroglyphs alphabet exceeds 10000 (while the alphabet for Latin-base languages is usually less than 100) the recognition of hieroglyphs introduces new challenges to both training and recognizing processes. One of the approaches to this task is to segment the characters into keys, recognize them and then compose the final answer [10]. In this case, the neural network with thousands of outputs is not required. Even though this method can solve the task it has drawbacks: non-robust to different image distortions and the high dependence on the quality of the segmentation. Some authors use very large (more than $10^7$ trainable parameters) and deep neural networks which demonstrates outstanding quality but requires huge computational power and cannot be used in mobile solutions [11, 12]. Another group of solutions deals only with the part of the alphabet [13]. In this case, some hieroglyph-specific problems fade and the recognition process can be build using classical approaches.

### 2.2 Embedded Nets

Embedded nets are commonly used instruments in the field of image recognition. The main idea of embedded nets is to map input data to point in embedded space where points of similar data must be closer to each other than points of different data; the more different the input data is, the greater the distance between their representations in embedded space should be. For training, the embedded nets contrastive loss (Siamese architecture) [14] or the triplet loss (Triplet architecture) [15] is used. Contrastive loss tries to maximize the distance between two samples from different classes and minimize the distance between samples from the same class. Triplet loss is more complex, it is trying to maximize the gap between distances for two samples of the same class and distance for samples from different classes, where one sample is used in both cases. Embedded nets are being used for recognition [16, 17], verification [18], object tracking [19]. The main advantage of this approach is that the size of the resulting neural network does not directly depend on the size of the alphabet. This makes them very perspective in tasks with a large number of classes, for instance, in hieroglyph recognition.

## 3. DATASET AND EVALUATION

We evaluate our method on the PHD08 dataset [20]. Which has images of Korean hieroglyphs of 2350 classes (2187 per class, nine different fonts). A total of 5139450 binary images of hieroglyphs, all samples having a different size, rotation, and distortions. An example of the final grayscaled Korean character input data is shown in Fig. 4.

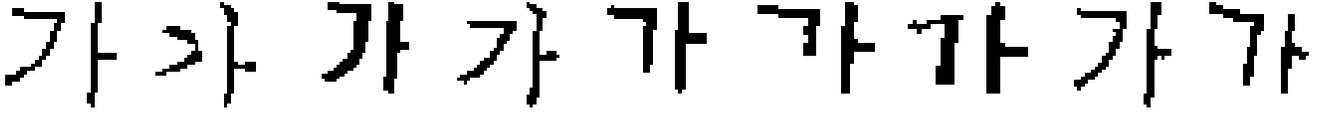

Figure 4. Example of images of Korean symbol U+AC00 from PHD08 dataset

The accuracy($Acc$) for classification was defined as:

$$Acc = \frac{\sum_i^N [c_i == y_i]}{N} * 100\% \qquad (1)$$

Where $c_i$ is the output label from classifier for image $i$, $y_i$ is the ground truth output label for image $i$, $N$ - total number of images in the dataset. $[c_i == y_i]$ is the indicator function which returns 1 if the statement is true and 0 otherwise.

## 4. SIAMESE TRAINING METHOD

The Siamese training method was first introduced in 1990 to verify the signature in the context of the problem of image matching [21]. For the training, process architecture consists of two twin networks, which receive different data at the input, but whose outputs are connected by a layer of the distance function. This function finds the distance between the representations of the last layers of networks. In this case, the configurable network parameters are the same for both branches. This system has two key properties:

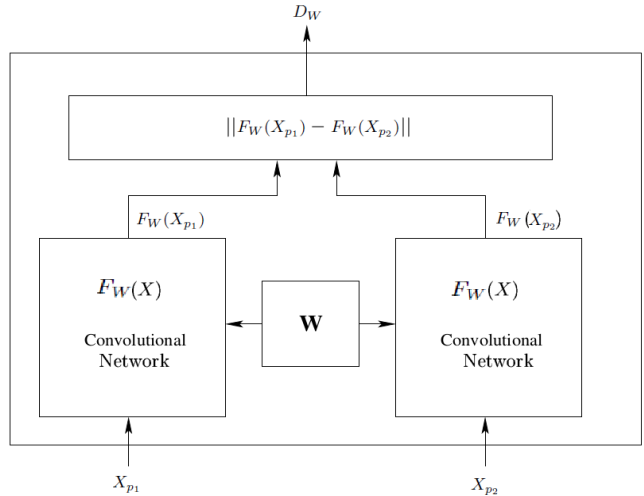

Figure 5. The training architecture.

• Consistency of representations: since the parameters of two neural networks are communicated, the representations of a pair of very similar images cannot be very different.

• Symmetry: The answer that such a network will give to a pair of images will be the same regardless of which image to which of the input layer is applied.

The training architecture shown In Fig 5.

## 5. USING EMBEDDED NET FOR CLASSIFICATION

After training embedded net we use training data to calculate representation of a reference sample of class for every hieroglyph from the alphabet:

$$d_{y_i} = P(x_i); \forall y_i \in Y \quad (2)$$

$$d_{y_i} \in D \quad (3)$$

Where $D$ is a plurality of referent representation of symbols, $Y$ is a plurality of symbols, $x_i$ is $i$ - th image of hieroglyph from training data which associates with $y_i$ symbol, and $P(*)$ is an operator of applying of embedded net. The reference representation is obtained from the image of the character without any additions. Then using the method of nearest neighbor we consider that the answer of the classifier is the symbol, whose reference representation will be closer to the representation of the image received at the input:

$$y *= argmin_{y \in Y}(p(P(x *), d_y)) \quad (4)$$

Where $y *$ is an answer of the classifier, $x *$ is an input image of hieroglyph, $p(*,*)$ is a metric in embedded space.

## 6. NET TRAINING

### 6.1 Train Data

Synthetically generated data was used to train the network [22]. For this, a background image was selected, on which hieroglyphs were subsequently applied. For training, the entire alphabet of Korean hieroglyphs (11172) was used to make a correct comparison with other recognition systems. For testing, it was used as an open dataset that has only 2350 characters from the Korean alphabet.

The next step is to cut out only the images of hieroglyphs from the images which were used, transform them to grayscale, and sort by class. For better learning, the images of hieroglyphs are also augmented with projective transformations and rotations [23]. In the end, there are pairs formed from the received images on which the net is being trained.

### 6.2 Architecture of Embedded Convolution Neural Network

For the embedded net we have chosen architecture which is described in Table 1.

Table 1. Architecture of embedded net

| Layers | | | |
|---|---|---|---|
| # | Type | Parameters | Activation function |
| 1 | conv | 16 filters $3 \times 3$, stride $1 \times 1$, no padding | softsign |
| 2 | conv | 16 filters $5 \times 5$, stride $2 \times 2$, padding $2 \times 2$ | softsign |
| 3 | conv | 16 filters $3 \times 3$, stride $1 \times 1$, padding $1 \times 1$ | softsign |
| 4 | conv | 24 filters $5 \times 5$, stride $2 \times 2$, padding $2 \times 2$ | softsign |
| 5 | conv | 24 filters $3 \times 3$, stride $1 \times 1$, padding $1 \times 1$ | softsign |
| 6 | conv | 24 filters $3 \times 3$, stride $1 \times 1$, padding $1 \times 1$ | softsign |
| 7 | fc | 25 outputs | |

$$softsign(x) = \frac{x}{1+|x|} \quad (5)$$

The contrastive loss function[24]. $l_2$ norm was used for training as the distance in the embedded space. The input image size was $37 \times 37$, grayscale. The proposed architecture has only $7.76 \times 10^4$ weights. For comparison, the classification net of the same architecture for the full alphabet will have $2.18 \times 10^7$ weights.

### 6.3 Pairs matching

In the initial experiment, we have generated all pairs randomly regardless of the presented symbol. The only restriction is that we did not allow the reduplication of the pairs. We have also used image augmentation only for the first image in the pair in the training process. Pair samples for the training are presented in Fig 6.

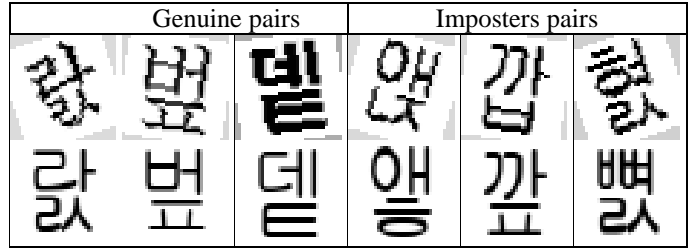
Figure 6. Examples of net training images.

### 6.4 Training

The network was trained on pairs of images of hieroglyphs. There was a total of $1452360$ pairs (for every class of hieroglyph was chosen $130$ pairs of imposters and $65$ pairs of genuine), while the number of image pairs belonging to one class was equal to the number of image pairs belonging to different classes. The data were randomly divided into training and validation samples in a ratio of $4:1$. The network was trained for $500$ epochs. The stochastic gradient descent method was used as an optimizer.

Trained embedded net shows $\mathbf{48.84\%}$ of $Acc$. To compare the results we use the recently released version of Tesseract 4.0.0 [9, 25], which reaches $37.67\%$ of $Acc$. Comparing to Tesseract 4.0.0 our performance is better but still is not sufficiently good. Important to note that the classification neural network of such an architecture did not converge with default parameters. We hypothesize that this is due to the enormously large size of the last fully connected layer but this question needs further research which is not the subject of the current work.

### 6.5 Pairs Matching Using Key Symbols

While analyzing the errors, we have found out that a neural network typically cannot distinguish similar hieroglyphs. To address this problem and improve the classification quality we suggest to change the pairs generation process. Since Korean hieroglyphs are combined from specific keys we can generate imposters using similar hieroglyphs that share some of their keys. To study the amount of the shared keys we conducted two additional experiments.

In the first experiment, all imposters were sharing two keys. In this case, the neural network tries to find differences between similar hieroglyphs and the quality should be increased according to the result of the error analysis. Fig. 7 illustrates different hieroglyphs which are sharing two keys. On the contrary, in the second experiment, we have created an imposter pair from the hieroglyphs with zero shared keys to study the dependency between similarity in the training data and classification quality.

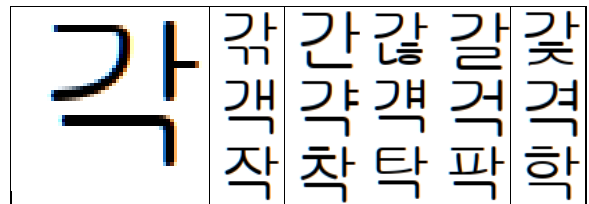
Figure 7. Example of variability of keys for korean hieroglyphs.

### 7. RESULTS

Results of the initial and two additional experiments are presented in Table 2. The best quality is $60.83\%$ which is achieved using similar hieroglyphs in imposter pairs. From the result, we can see that introducing similar but different data is crucial and affects quality very much. In its best form, our method generates more than $1.5$ times fewer errors than an open OCR engine Tesseract. The important property of the suggested method is that it allows us to train relatively small neural networks that are possible to use on mobile devices and in other situations where computational power is limited.

Table 2. Results

|  | Embedded net | | | Tesseract OCR 4.0.0 |
|---|---|---|---|---|
|  | 0 keys | 2 keys | random |  |
| $Acc$ | 26.78 % | **60.83 %** | 48.84 % | 37.67 % |

## 8. CONCLUSION

In this work, we have proposed a method for the embedded neural network training for Korean hieroglyphs recognition. Our study shows that the quality highly depends not only on the presence of different symbols in the training data, but also on the similarity of the imposter pairs. The neural network, trained with the suggested method, has a very small number of weights in comparison to the heavy state-of-the-art methods that require huge computational power. On the contrary, the neural network trained with the suggested method can be used on mobile devices and in other special cases with limited computational power. Still, in comparison with the Tesseract open-source OCR engine, our neural network shows a very good quality.

In our future works, we plan to study this field further and generalize the pair generation method for different languages and will try to rely not on the hieroglyph keys but the training process since in some tasks it is impossible to obtain them. We will use the trained net answers instead of the information from the keys of hieroglyphs, to extend and improve our method for not systematic data. Along with that, we plan to improve the quality by using more complex final space and loss function.

## ACKNOWLEDGMENTS

This work is partially supported by the Russian Foundation for Basic Research (projects 17-29-03370, 17-29-07093).